\documentclass{article}

    \PassOptionsToPackage{numbers, compress}{natbib}

    \usepackage[preprint]{neurips_2022}

\usepackage[utf8]{inputenc} 
\usepackage[T1]{fontenc}    
\usepackage{hyperref}       
\usepackage{url}            
\usepackage{booktabs}       
\usepackage{amsfonts}       
\usepackage{nicefrac}       
\usepackage{microtype}      
\usepackage{xcolor}         
\usepackage{changepage} 
\usepackage{adjustbox}
\usepackage{subcaption} 
\usepackage{wrapfig}
\usepackage{siunitx}
\usepackage{enumitem}
\usepackage{amsmath,amsfonts,amssymb}
\usepackage{cleveref}
\usepackage{tabularx}
\usepackage{xspace}
\usepackage{xltabular}
\usepackage{multirow}
\usepackage{caption}

\usepackage{amsmath,amsfonts,bm}

\def\eqref#1{equation~\ref{#1}}

\def\1{\bm{1}}

\def\vx{{\bm{x}}}
\def\vy{{\bm{y}}}

\DeclareMathAlphabet{\mathsfit}{\encodingdefault}{\sfdefault}{m}{sl}
\SetMathAlphabet{\mathsfit}{bold}{\encodingdefault}{\sfdefault}{bx}{n}

\def\gD{{\mathcal{D}}}

\def\gT{{\mathcal{T}}}

\newcommand{\teacher}{\boldsymbol{\theta^{\text{T}}}}
\newcommand{\student}{\boldsymbol{\theta^{\text{S}}}}
\newcommand{\taskdata}{\gD^{\gT}}
\newcommand{\augdata}{\widetilde{\gD}^{\gT}}

\definecolor{gred}{RGB}{219,68,55}
\definecolor{gblue}{RGB}{66,133,244}
\definecolor{gyellow}{RGB}{244,180,0}
\definecolor{ggreen}{RGB}{15,157,88}
\definecolor{ggrey}{RGB}{115,115,115}

\newcommand{\colorG}[1]{\textcolor{ggreen}{\textbf{#1}}}

\newcommand{\method}{Ours\xspace}

\title{Synthetic Data Generation in Low-Resource Settings \\via Fine-Tuning of Large Language Models}

\author{%
Jean Kaddour \\
University College London \\
  \And
  Qi Liu \\
  University of Hong Kong \\
}

\begin{document}

\maketitle

\begin{abstract}
The in-context learning ability of large language models (LLMs) enables them to generalize to novel downstream tasks with relatively few labeled examples. However, they require enormous computational resources to be deployed. Alternatively, smaller models can solve specific tasks if fine-tuned with enough labeled examples. These examples, however, are expensive to obtain. In pursuit of the best of both worlds, we study synthetic data generation of fine-tuning training data via fine-tuned teacher LLMs to improve the downstream performance of much smaller models. In four text classification and two text generation tasks, we find that both data generation and annotation dramatically improve the respective downstream model's performance, occasionally necessitating only a minor fraction of the original training dataset.
\end{abstract}

\section{Introduction}
Large language models (LLMs) have demonstrated \emph{in-context learning} (ICL) capabilities in various natural language processing tasks, which allow us to perform an unseen downstream task by prompting the model with a collection of input-target pairs and a single unlabeled example \cite{gpt3}. Crucially, ICL requires relatively few labeled examples but large model sizes \cite{wei2022emergent}. However, deploying LLMs in real-world systems is challenging due to their computational costs and inference latency \cite{callms}. 

An alternative paradigm that enables good results with much smaller models is to specialize a pre-trained model for a single task through gradient-based \emph{supervised fine-tuning} (SFT) \cite{finetuning_1, bert}. The drawback of this approach is that it relies on labeled examples, which require human annotators and, therefore, is expensive and time-consuming. Especially in low-resource settings with only a handful of examples, SFT can be challenging \cite{zhangRevisitingFewsampleBERT2021b}.

In this work, we attempt to yield the best of both worlds by fine-tuning smaller models with training data generated by an LLM. Following recent work on training data generation \cite{dino,supergen,callms}, we show that by (i) \emph{annotating} unlabeled examples or (ii) \emph{generating} entirely new ones, we can effectively transfer knowledge from the LLM (\emph{teacher}) to the specialized model (\emph{student}), which can be several magnitudes of orders smaller, akin to knowledge distillation \cite{hinton2015distilling} but only via the exchange of data.

We find that synthetic data generation after fine-tuning the teacher LLM, even on only extremely limited data, improves the synthetic data quality as measured by the downstream model generalization performance. For example, fine-tuning a 20B LLM on as few as 125 examples (5\% of the RTE dataset \cite{superglue}) increases the augmented-data-fine-tuned downstream model's performance by multiple percentage points. 

We empirically verify our approach on four text classification and two natural language generation tasks, finding that both (i) and (ii) consistently improve the student model's downstream performance. We provide ablation studies on varying amounts of synthetic data, comparisons with GPT3.5 \cite{gpt3} as a teacher model, and evaluate the teacher LLMs directly on the downstream tasks. 

\begin{figure}[t]
    \centering
    \includegraphics[width=\columnwidth]{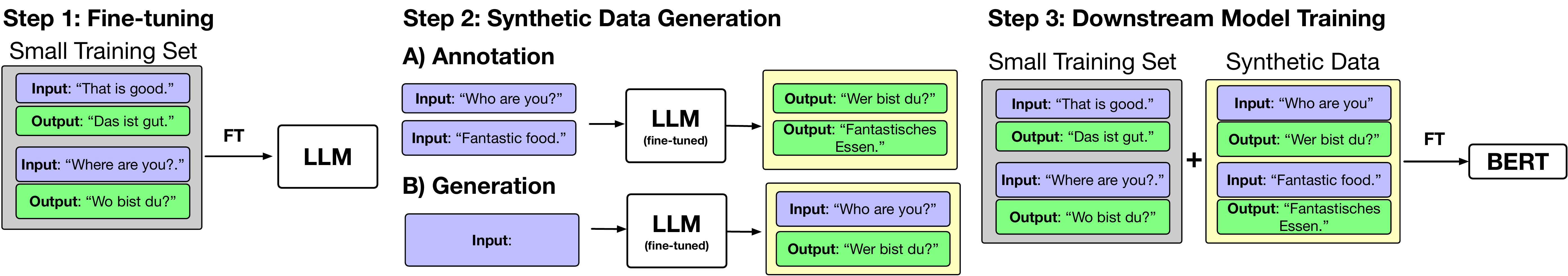}
    \caption{\textbf{Synthetic Data Generation via Fine-Tuning the Teacher LLM}: \emph{Step 1:} We fine-tune (FT) an LLM (GPT-NeoX-20B \cite{neox}). \emph{Step 2:} We either annotate unlabeled instances or generate entirely new ones. \emph{Step 3:} We train a small downstream model on the augmented training dataset.}
    \label{fig:overview}
\end{figure}

\section{Method}
\subsection{Problem Definition}
Consider a task-specific and labeled dataset $\taskdata:= \{(\vx_i, \vy_i) \}_{i=1}^N$ of text input-output pairs for some task $\gT$, a teacher LLM paratemerized by $\teacher$ and lightweight student model $\student$, with $|\student| \ll |\teacher|$. We aim to yield $\student$ that performs similarly to $\teacher$ in solving $\gT$. The premise is that $\teacher$ performs much better than $\student$ but is more expensive to deploy. Our strategy is to generate synthetic data $\augdata$ using $\teacher$, and then fine-tune $\student$ on $\gD^{\gT} \cup \augdata$.

We further distinguish the cases of (i) \emph{annotating} an unlabeled set of instances $\{ \vx_i \}_{i=1}^M$ such that $\augdata := \{ (\vx_i, \hat \vy_i)\}_{i=1}^M$, or (ii) \emph{generating} entire input-output pairs $\augdata := \{ (\hat \vx_i, \hat \vy_i)\}_{i=1}^M$, where $\hat \vx_i, \hat \vy_i$ refer to synthetic input and output sequences, respectively. (i) is relevant in scenarios where abundant unlabeled text is available (e.g., scraped from the internet), but obtaining annotations is expensive. In turn, (ii) is suitable even in situations without unlabeled instances. 

\subsection{Teacher model} The teacher LLM is the 20B GPT-NeoX \cite{neox}, which, at the time of our experimentation phase, was considered state-of-the-art \footnote{In the meantime, many better-performing models have been released \cite{callms,OpenLLMLeaderboarda}, which would likely further improve the effectiveness of our approach.}. We adopt the text-to-text framework \cite{T5}, where the LLM receives a text query as input and produces text output. For instance, when generating dialog responses, the query corresponds to incomplete dialog utterances, and the model generates the response. As for classification tasks, the query corresponds to the textual input, and the model generates the class label. 

Further, we fine-tune GPT-NeoX for multi-task instruction-following on several task mixtures, using the Super-NaturalInstructions data set \cite{wang2022supernaturalinstructions}. Each prompt consists of a task description prefix and a small subset of training input-output pairs, separated by \texttt{[INPUT]} and \texttt{[OUTPUT]} tags. Further, depending on whether we annotate unlabeled text or generate entirely new examples, the prompt ends with either \texttt{[OUTPUT]} or \texttt{[INPUT]}, respectively.

We fine-tune all layers and use Adam for up to $5$ epochs with a cosine schedule and a learning rate of $1e-5$. Our code relies upon the GPT-Neox library \cite{Andonian_GPT-NeoX_Large_Scale_2021}, which utilizes DeepSpeed \cite{rasley2020deepspeed} to facilitate distributed training. We use NVIDIA A100 GPUs for all experiments as provided by an internal compute cluster. 

An important inference hyper-parameter is the generation temperature, which controls the LLM output randomness. We find that a low temperature of $0.1$ works best for annotations, while a high temperature of $0.8$ works best for generations (due to higher diversity in the samples).  

\subsection{Student models}
For text classification and generation tasks, we use RoBERTa-Large \cite{roberta}, and BART-Large \cite{bart} models, respectively. We fine-tune their publicly available checkpoints. We fine-tune up to $320$ epochs for TC tasks using Adam, a batch size of $50$, and a learning rate of $1e-5$. We fine-tune with Adam for $5$ epochs for TG tasks, using a linear learning schedule, batch size $32$, and learning rate $5e-5$. We report the results using the model with the best validation loss.

\section{Experiments}
\subsection{Datasets}
\begin{wraptable}{r}{4cm}
\centering
\resizebox{4cm}{!}{%
\begin{tabular}{@{}c|ccc@{}}
\toprule
\multirow{2}{*}{\textbf{Dataset}} & \multicolumn{3}{c}{\textbf{\# Examples}} \\ \cmidrule(l){2-4} 
                         & \textbf{Train}      & \textbf{Dev}     & \textbf{Test}     \\ \midrule
SLURP \cite{slurp}                      & 11514       & 2033     & 2974     \\
RTE \cite{superglue}                     & 2500       & 277     & 300     \\
BoolQ \cite{boolq}                      & 9427       & 3270     & 3245 \\
MultiRC \cite{multirc}                      & 5100       & 953     & 1800     \\
PubMedQA \cite{jin-etal-2019-pubmedqa}                      & 212300       & -     & 1000     \\ \midrule
SGD \cite{sgd}                      & 164982       & 10000     & 10000     \\
WebNLG \cite{webnlg}                      & 35426       & 1667     & 1779     \\
 \bottomrule
\end{tabular}%
}
\caption{\textbf{Dataset Statistics.}}
\label{tab:statistics}
\vspace{-5ex}
\end{wraptable}
We list the dataset sizes in \Cref{tab:statistics}. One can apply the fractions in later tables to this table to calculate the number of fine-tuning samples for the teacher LLM. 
\paragraph{Text classification}
We report classification accuracies across four classification tasks. \emph{SLURP} (Spoken Language Understanding Resource Package) \cite{slurp} is a multi-domain dataset for end-to-end spoken language understanding, consisting of single-turn user interactions with a home assistant. 
\emph{BoolQ} \cite{boolq} contains naturally occurring yes/no questions about a provided paragraph. 
\emph{RTE} (Recognizing Textual Entailment) \cite{glue} is a dataset for recognizing whether, given two texts, the meaning of one is entailed (can be inferred) from the other.
\emph{MultiRC} (Multi-Sentence Reading Comprehension) \cite{multirc} is a multi-sentence question-answering dataset, where each input consists of a short paragraph and question, which requires combining information from multiple sentences. 

\paragraph{Natural language generation}

We report Rouge-\{1,2,L\} scores \cite{lin2004rouge} on both development and test sets across two tasks. \emph{Schema Guided Dialog} (SGD) \cite{sgd} includes task-oriented conversations between a human and a virtual assistant. Here, we follow \citet{gem} and construct the task of generating an utterance given an unfinished dialog as input. \emph{WebNLG} is a data-to-text \cite{webnlg} task, which aims at generating a text that verbalizes the input triples grammatically correct.

\subsection{Main results}
\label{sec:main_results}
\begin{figure}
\begin{minipage}[b]{.5\linewidth}
\centering
\resizebox{5cm}{!}{%
\begin{tabular}{@{}cccccc@{}}
\toprule
\multirow{2}{*}{\textbf{Dataset}} &
  \multirow{2}{*}{\textbf{Type}} &
  \multicolumn{2}{c}{\textbf{Data amount in \%}} &
  \multirow{2}{*}{\textbf{Dev Acc}} &
  \multirow{2}{*}{\textbf{Test Acc}} \\ \cmidrule(lr){3-4}
                                              &       & \textbf{Original}     & \textbf{\method} &       &                      \\ \midrule
\multicolumn{1}{c|}{\multirow{10}{*}{\shortstack{SLURP \\\cite{slurp}}}}   &       & \multirow{3}{*}{1\%}  & 0 \%                 & 42.25 & 43.95                \\
\multicolumn{1}{c|}{}                         & $X, Y$  &                       & 26   \%              & 54.57 & 54.25                \\
\multicolumn{1}{c|}{}                         & $Y | X$ &                       & 78  \%               & \textbf{76.14} & \textbf{76.09}                \\ \cmidrule(l){2-6} 
\multicolumn{1}{c|}{}                         &       & \multirow{3}{*}{5\%}  & 0 \%                 & 73.49 & 71.59                \\
\multicolumn{1}{c|}{}                         & $X, Y$  &                       & 43 \%          & 77.39 & 76.89                \\
\multicolumn{1}{c|}{}                         & $Y | X$ &                       & 78 \%                & \textbf{85.00} & \textbf{83.96}                \\ \cmidrule(l){2-6} 
\multicolumn{1}{c|}{}                         & -     & \multirow{3}{*}{10\%} & 0\%                  & 80.12 & 80.04                \\
\multicolumn{1}{c|}{}                         & $X, Y$  &                       & 43  \%               & 82.59 & 81.91                \\
\multicolumn{1}{c|}{}                         & $Y | X$ &                       & 43 \%                & \textbf{86.13} & \textbf{86.48}                \\ \cmidrule(l){2-6} 
 \multicolumn{1}{c|}{} & -     & 100 \%                 & 0  \%                  & 88.64 & 87.70  \\  \midrule
\multicolumn{1}{c|}{\multirow{10}{*}{\shortstack{BoolQ \\\cite{boolq}}}}   &       & \multirow{3}{*}{1\%}  & 0    \%              & 62.84 & \multirow{10}{*}{N/A} \\
\multicolumn{1}{c|}{}                         & $X, Y$  &                       & 31 \%                & 68.96 &                      \\
\multicolumn{1}{c|}{}                         & $Y | X$ &                       & 44   \%             & \textbf{79.72} &                      \\ \cmidrule(lr){2-5}
\multicolumn{1}{c|}{}                         &       & \multirow{3}{*}{5\%}  & 0    \%              & 62.97 &                      \\
\multicolumn{1}{c|}{}                         & $X, Y$  &                       & 31 \%                & 66.02 &                      \\
\multicolumn{1}{c|}{}                         & $Y | X$ &                       & 44 \%               & \textbf{80.09} &                      \\ \cmidrule(lr){2-5}
\multicolumn{1}{c|}{}                         & -     & \multirow{3}{*}{10\%} & 0    \%              & 68.29 &                      \\
\multicolumn{1}{c|}{}                         & $X, Y$  &                       & 31  \%               & 77.22 &                      \\
\multicolumn{1}{c|}{}                         & $Y | X$ &                       & 44   \%              & \textbf{81.93} &                      \\  \cmidrule(l){2-5} \multicolumn{1}{c|}{} & -     & 100 \%  & 0  \%  & 85.05 &   \\ \midrule
\multicolumn{1}{c|}{\multirow{10}{*}{\shortstack{RTE \\\cite{glue}}}}     &       & \multirow{3}{*}{5\%}  & 0     \%             & 60.65 & \multirow{10}{*}{N/A} \\
\multicolumn{1}{c|}{}                         & $X, Y$  &                       & 80   \%              & 66.79 &                      \\
\multicolumn{1}{c|}{}                         & $Y | X$ &                       & 80  \%               & \textbf{83.20} &                      \\ \cmidrule(lr){2-5}
\multicolumn{1}{c|}{}                         &       & \multirow{3}{*}{10\%} & 0  \%                & 65.43 &                      \\
\multicolumn{1}{c|}{}                         & $X, Y$  &                       & 80   \%              & 69.68 &                      \\
\multicolumn{1}{c|}{}                         & $Y | X$ &                       & 80  \%               & \textbf{83.75} &                      \\ \cmidrule(lr){2-5}
\multicolumn{1}{c|}{}                         & -     & \multirow{3}{*}{20\%} & 0  \%                & 74.73 &                      \\
\multicolumn{1}{c|}{}                         & $X, Y$  &                       & 80   \%              & 76.84 &                      \\
\multicolumn{1}{c|}{}                         & $Y | X$ &                       & 80  \%               & \textbf{85.20} &  \\ \cmidrule(l){2-5} 
 \multicolumn{1}{c|}{} & -     & 100 \%                 & 0  \%                  &  86.60  \\ \midrule
\multicolumn{1}{c|}{\multirow{10}{*}{\shortstack{MultiRC \\ \cite{multirc}}}} &       & \multirow{3}{*}{1\%}  & 0 \%                 & 57.50 & 8.18                 \\
\multicolumn{1}{c|}{}                         & $X, Y$  &                       & 40  \%               & 63.40 & 15.11                \\
\multicolumn{1}{c|}{}                         & $Y | X$ &                       & 254  \%              & \textbf{71.46} & \textbf{24.24}                \\ \cmidrule(l){2-6} 
\multicolumn{1}{c|}{}                         &       & \multirow{3}{*}{5\%}  & 0   \%               & 67.70 & 18.05                \\
\multicolumn{1}{c|}{}                         & $X, Y$  &                       & 40  \%               & \textbf{72.85} & 21.86                \\
\multicolumn{1}{c|}{}                         & $Y | X$ &                       & 254  \%             & 71.51 & \textbf{32.63}                \\ \cmidrule(l){2-6} 
\multicolumn{1}{c|}{}                         & -     & \multirow{3}{*}{10\%} & 0   \%               & 70.63 & 22.35                \\
\multicolumn{1}{c|}{}                         & $X, Y$  &                       & 40 \%                & 73.88 & 24.97                \\
\multicolumn{1}{c|}{}                         & $Y | X$ &                       & 254  \%              & \textbf{76.90} & \textbf{34.94}               
 \\ \cmidrule(l){2-6} 
 \multicolumn{1}{c|}{} & -     & 100\%                 & 0 \%                  &  82.12 & 48.16  \\ \bottomrule
\end{tabular}%
}
\caption{Text Classification.}
\end{minipage}
\begin{minipage}[b]{.5\linewidth}
\centering
\resizebox{8cm}{!}{\begin{tabular}{@{}c|c|cc|cccccc@{}}
\toprule
\multirow{2}{*}{\textbf{Dataset}} &
  \multirow{2}{*}{\textbf{Type}} &
  \multicolumn{2}{c|}{\textbf{Data amount in \%}} &
  \multicolumn{3}{c}{\textbf{Validation scores}} &
  \multicolumn{3}{c}{\textbf{Test scores}} \\ \cmidrule(l){3-10} 
 &
   &
  \textbf{Original} &
  \textbf{\method} &
  \textbf{R-1} &
  \textbf{R-2} &
  \textbf{R-L} &
  \textbf{R-1} &
  \textbf{R-2} &
  \textbf{R-L} \\ \midrule
\multirow{10}{*}{\begin{tabular}[c]{@{}c@{}}\shortstack{SGD \\\cite{sgd}}\end{tabular}} &
  - &
  \multirow{3}{*}{1\%} &
  0 \% &
  21.78 &
  9.29 &
  20.09 &
  21.91 &
  9.40 &
  20.06 \\
 & $X, Y$  &                       & \multirow{2}{*}{10\%} & 43.72 & 25.09 & 39.84 & 40.82 & 22.36 & 37.00 \\
 & $Y | X$ &                       &                      & \textbf{47.44} & \textbf{28.19} & \textbf{43.47} & \textbf{43.37} & \textbf{24.49} & \textbf{39.58} \\ \cmidrule(l){2-10} 
 & -     & \multirow{3}{*}{5\%}  & 0   \%                 & 33.17 & 16.86 & 30.32 & 30.92 & 14.86 & 28.09 \\
 & $X, Y$  &                       & \multirow{2}{*}{10\%}  & 45.27 & 26.39 & 41.48 & 41.71 & 23.25 & 37.98 \\
 & $Y | X$ &                       &                      & \textbf{48.99} & \textbf{29.87} & \textbf{45.02} & \textbf{44.60} & \textbf{25.76} & \textbf{40.66} \\ \cmidrule(l){2-10} 
 & -     & \multirow{3}{*}{10\%} & 0   \%                 & 35.48 & 18.48 & 32.47 & 33.40 & 16.40 & 30.37 \\
 & $X, Y$  &                       & \multirow{2}{*}{10\%}  & 48.49 & 29.59 & 44.62 & 43.89 & 25.29 & 40.16 \\
 & $Y | X$ &                       &                      & \textbf{49.31} & \textbf{30.15} & \textbf{45.32} & \textbf{44.89} & \textbf{25.94} & \textbf{40.98} \\ \cmidrule(l){2-10} 
 & -     & 100\%                 & 0   \%                 & 57.62 & 39.64 & 53.63 & 50.23 & 32.03 & 46.30 \\ \midrule
\multicolumn{1}{c|}{\multirow{10}{*}{\shortstack{WebNLG \\ \cite{webnlg}}}} &
  - &
  \multirow{3}{*}{1\%} &
  0 \%&
  53.94 &
  31.76 &
  42.58 &
  50.21 &
  28.44 &
  39.72 \\
 & $X, Y$  &                       & \multirow{2}{*}{10\%} & 72.99 & 47.16 & 56.06 & 69.63 & 42.99 & 52.98 \\
 & $Y | X$ &                       &                      & \textbf{75.69} & \textbf{51.18} & \textbf{59.60} & \textbf{70.21} & \textbf{44.34} & \textbf{54.51} \\ \cmidrule(l){2-10} 
 & -     & \multirow{3}{*}{5\%}  & 0 \%                   & 56.58 & 35.74 & 46.46 & 53.17 & 32.57 & 44.65 \\
 & $X, Y$  &                       & \multirow{2}{*}{10\%}  & 76.57 & 51.23 & 60.12 & 71.63 & 44.38 & 54.37 \\
 & $Y | X$ &                       &                      & \textbf{78.02} & \textbf{53.60} & \textbf{61.46} & \textbf{72.11} & \textbf{45.50} & \textbf{55.29} \\ \cmidrule(l){2-10} 
 & -     & \multirow{3}{*}{10\%} & 0 \%                   & 59.02 & 38.21 & 48.37 & 54.07 & 32.47 & 44.06 \\
 & $X, Y$  &                       & \multirow{2}{*}{10\%}  & 77.72 & 53.56 & 61.60 & 72.02 & 45.20 & 54.71 \\
 & $Y | X$ &                       &                      & \textbf{78.38} & \textbf{54.56} & \textbf{62.72} & \textbf{72.73} & \textbf{46.02} & \textbf{55.96} \\ \cmidrule(l){2-10} 
 & -     & 100\%  & 0 \%                   & 80.65 & 58.50 & 65.73 & 73.27 & 46.90 & 55.76 \\ \bottomrule
\end{tabular}}
\caption{Natural Language Generation.}
\end{minipage}
\label{tab:classification}
\caption{\textbf{Comparison of Performances with and without \method.} We find that \method is most effective in settings with the smallest amount of initial data despite the teacher LLM being fine-tuned on that small amount of data. For reference, we list the performance of the student model when fine-tuned on 100\% training data.}
\end{figure}

\Cref{tab:classification} shows the main results. We calculate the data amounts relatively w.r.t. the original training dataset size. For example, an entry with 1\% original and 10\% synthetic data means we construct the augmented training dataset by concatenating 1\% of data randomly sampled from the original training dataset and 10\% synthetic data.

We observe that both data annotations and generations improve performance across all scores. Interestingly, the smaller the initial training dataset, the larger the gains. This is interesting because it implies that we do not need many samples for the teacher's fine-tuning phase to succeed.

\subsection{Varying amount of synthetic data}
To understand the relationship between the amount of synthetic data added and its downstream performance improvements, we vary the amount of synthetic data for the NLG tasks.

\begin{figure}

\begin{minipage}[b]{0.45\textwidth}
\centering
\resizebox{4.5cm}{!}{%
\begin{tabular}{@{}c|ccccc@{}}
\toprule
\multirow{2}{*}{\textbf{Dataset}} &
  \multirow{2}{*}{\textbf{Type}} &
  \multicolumn{2}{c}{\textbf{Data amount}} &
  \multicolumn{2}{c}{\textbf{Rouge-L}} \\ \cmidrule(l){3-6} 
 &
   &
  \textbf{Original} &
  \textbf{\method} &
  \textbf{Dev} &
  \textbf{Test} \\ \midrule
\multirow{8}{*}{SGD}     & -                      & 1\%                  & 0\%  & 20.09 & 20.06  \\ \cmidrule(l){2-6} 
                         & \multirow{3}{*}{$Y | X$} & \multirow{3}{*}{1\%} & 10\% & 43.47 & 39.58  \\
                         &                        &                      & 20\% & 45.22 & 40.89  \\
                         &                        &                      & 30\% & \textbf{46.02} & \textbf{41.29}  \\ \cmidrule(l){2-6} 
                         & \multirow{3}{*}{$X, Y$}  & \multirow{3}{*}{1\%} & 10\% & 39.84 & 37.00  \\
                         &                        &                      & 20\% & 40.92 & 37.62  \\
                         &                        &                      & 30\% & \textbf{42.27} & \textbf{39.30}  \\ \cmidrule(l){2-6} 
                         & -                      & 100\%                & 0\%  & 53.63 & 46.30  \\ \midrule
\multirow{12}{*}{WebNLG} & -                      & 1\%                  & 0\%  & 42.58 & 39.72 \\ \cmidrule(l){2-6} 
                         & \multirow{5}{*}{$Y | X$} & \multirow{5}{*}{1\%} & 1\%  & 57.59 & 52.92  \\
                         &                        &                      & 2\%  & 58.94 & \textbf{54.50}  \\
                         &                        &                      & 3\%  & 59.44 & 54.44  \\
                         &                        &                      & 4\%  & \textbf{59.79} & 54.42  \\
                         &                        &                      & 5\%  & 59.56 & 54.44  \\ \cmidrule(l){2-6} 
                         & \multirow{5}{*}{$X, Y$}  & \multirow{5}{*}{1\%} & 1\%  & \textbf{57.09} & \textbf{52.81}  \\
                         &                        &                      & 2\%  & 57.00 & 52.29  \\
                         &                        &                      & 3\%  & 57.08 & 52.64  \\
                         &                        &                      & 4\%  & 56.92 & 52.72  \\
                         &                        &                      & 5\%  & 56.92 & 52.76  \\ \cmidrule(l){2-6} 
                         & -                      & 100\%                & 0\%  & 65.73 & 55.76  \\ \bottomrule
\end{tabular}%
}
\caption{\textbf{Different amounts of \method data added to the student model training set.} While both the SGD and WebNLG tasks benefit from the synthetic data, we observe diminishing returns.}
\label{tab:ablation_amount}

\end{minipage} \hfill
\begin{minipage}[b]{0.45\textwidth}
\resizebox{7cm}{!}{
\begin{tabular}{@{}c|ccccc@{}}
\toprule
\multirow{2}{*}{\textbf{Dataset}} &
  \multirow{2}{*}{\textbf{Type}} &
  \multicolumn{2}{c}{\textbf{Data amount}} &
  \multicolumn{2}{c}{\textbf{Rouge-L}} \\ \cmidrule(l){3-6} 
 &
   &
  \textbf{Original} &
  \textbf{\method} &
  \textbf{Dev} &
  \textbf{Test} \\ \midrule
\multirow{4}{*}{SGD}     & -                      & 1\%                  & 0\%  & 20.09 & 20.06  \\
                         & $Y | X$ & 1\% & 10\% & \bf{43.47} & \bf{39.58}  \\ 
                         & $X, Y$  & 1\% & 10\% & 39.84 & 37.00  \\
                        & $Y | X$; $X, Y$   & 1\% & 5\% each & 42.46 & 39.18  \\
 \midrule
\multirow{4}{*}{WebNLG} & -                      & 1\%                  & 0\%  & 42.58 & 39.72 \\ 
                         & $Y | X$ & 1\% & 10\%  & \bf{59.60} &   \bf{54.51} \\
                         & $X, Y$  & 1\% & 1\%  & 56.06 & 52.98  \\
                          & $Y | X$; $X, Y$  & 1\% & 5\% each & 59.41 & 54.17  \\
 \bottomrule
\end{tabular}}
\caption{\textbf{Combining Annotation ($Y | X$) and Generation ($X, Y$)} equally performs almost as well as annotating the same amount, while the latter assumes access to unlabeled instances, which can be difficult in practice.}
\label{tab:combination}

\end{minipage}
\end{figure}
\Cref{tab:ablation_amount} shows the results of different augmentation amounts with 1\% original training data. For SGD, we observe that up until 30\%, the downstream model performance increases. However, for WebNLG, we observe diminishing returns and even slightly worse performance as the amount of synthetic data increases.

\subsection{Evaluating the teacher LLMs directly}
We report the performances of the teacher models, i.e., without generating additional data points and fine-tuning a student model, in \Cref{tab:eval_teacher_llm}. Interestingly, the fine-tuned NeoX 20B model outperforms the fine-tuned \texttt{davinci-002} model with 175B parameters (fine-tuned with default hyper-parameters as provided by OpenAI's API).

\subsection{Comparing Teacher LLMs NeoX-20B and GPT-3.5 175B}
We investigate how effective our framework is when combined with other teacher models. We compare NeoX-20B with OpenAI's GPT-3.5 175B \cite{gpt3} (called \texttt{davinci-002} in their API) in \Cref{fig:comparison_teacher_llms}. The fine-tuned NeoX model consistently produces better training data as measured in the student LM's downstream performance on the tasks' test sets. This is an interesting result, given the difference in model size.   

\begin{figure}
\centering
\begin{tabular}{@{}c|ccc|ccc@{}}
\toprule
\multirow{2}{*}{\textbf{Model}} & \multicolumn{3}{c|}{\textbf{Dev}} & \multicolumn{3}{c}{\textbf{Test}} \\ \cmidrule(l){2-7} 
                                & Rouge 1   & Rouge 2   & Rouge L   & Rouge 1   & Rouge 2   & Rouge L   \\ \midrule
davinci-002                     & 35.47     & 17.99     & 32.35     & 34.62     & 17.19     & 31.46     \\
NeoX                            & \bf 47.44     & \bf 28.19     & \bf 43.47     & \bf 43.37     & \bf 24.49     & \bf 39.59     \\ \bottomrule
\end{tabular}
\caption{\textbf{Comparing the Teacher LLMs}: We report the student LM's downstream performance after being fine-tuned on 11\% of the original training dataset size, where 1\% is the original data (also used to fine-tune the teacher LLMs) and 10\% teacher-annotated examples. The task is SGD.}
\label{fig:comparison_teacher_llms}
\end{figure}

\begin{figure}
\centering
\begin{tabular}{@{}c|cccccc@{}}
\toprule
\multirow{2}{*}{\textbf{Model}} & \multicolumn{3}{c}{\textbf{Dev}}                 & \multicolumn{3}{c}{\textbf{Test}} \\ \cmidrule(l){2-7} 
                                & Rouge 1 & Rouge 2 & \multicolumn{1}{c|}{Rouge L} & Rouge 1   & Rouge 2   & Rouge L   \\ \midrule
davinci-002                     & 30.71   & 14.38   & \multicolumn{1}{c|}{28.06}   & 30.54     & 14.14     & 27.70     \\
NeoX                            & \bf{38.38}   & \bf{20.84}   & \multicolumn{1}{c|}{\bf{35.44}}   & \bf{37.78}     & \bf{20.13}     & \bf{34.58}     \\ \bottomrule
\end{tabular}
\caption{\textbf{Evaluating the Teacher LLMs} on SGD after being fine-tuned with 1\% training data.}
\label{tab:eval_teacher_llm}
\end{figure}

\subsection{Combining annotation and generation}
In \Cref{sec:main_results}, we observe that annotations of existing, unlabeled data yield more significant gains than generating entire input-output pairs. In many real-world settings, however, access to unlabeled data may still be more costly than generating data points entirely de novo. Hence, we want to investigate whether adding generated data points to annotated ones in an equal amount can yield performances closer to a larger set of annotated ones while potentially being much cheaper in practice.

In the setting of 1\% original training data, we confirm this hypothesis in the affirmative, as shown in \Cref{tab:combination}. When mixing both data sources with 5\% each, we yield almost the same performance as if we had annotated 10\% of the whole data set. 

\subsection{Qualitative Analysis of Generated Text}
In \Cref{app:examples}, we showcase both annotated instances as well as entirely generated ones. By manually inspecting and verifying them, we find all generated examples coherent and syntactically correct. However, not all instances are factually truthful, a phenomenon often referred to as ``hallucination''. A post-hoc factuality filtering process of the generated examples could be an exciting avenue for future work \cite{li2023textbooks}.

\section{Related Work}
There is extensive literature on synthetic data augmentation for text data leveraging large teacher models \cite{bayer2022survey,callms}. However, most of these works do not fine-tune the teacher LLM on limited data to improve the quality of the generated data.
\citet{efrat2020turking} examine a model's ability to follow natural language instructions, including annotating unlabeled dataset examples. \citet{dino} study synthetic data generation on semantic textual similarity datasets without task-specific training examples. Similarly, other work has focused on information retrieval \cite{InPars}, code generation \cite{code_generation_knowledge_transfer}, and reasoning tasks \cite{ho2022large}. 

\citet{yoo2021gpt3mix} propose to transfer knowledge from LLMs to student models by generating synthetic examples and knowledge distillation using soft labels. \citet{reduce_label_cost} explore GPT-3 as a low-cost data labeler to train other models. For NLG and NLU tasks, they find that it costs 50\% to 96\% less to use GPT-3 generated labels than human annotations. Similarly, \citet{good_annotator} evaluates GPT-3's effectiveness as a data annotator on classification and named entity recognition tasks. 

\citet{chen2022weakly,zheng2023augesc,gunasekar2023textbooks,li2023textbooks} follow a similar approach to ours, augmenting pre-training data with synthetically generated documents. For example, \citet{gunasekar2023textbooks} generate textbooks and exercises to train a comparatively small model, which outperforms much bigger ones on coding tasks. In contrast, we consider the fine-tuning phase and classification and natural language generation tasks, similar to the work by \citet{sahu2022data,chen2022weakly} but with different tasks.

In spirit, our approach is similar to \emph{knowledge distillation} (KD) \cite{hinton2015distilling}, where one uses the teacher model's output logits as targets when training the student model. However, state-of-the-art LLMs are often served via cloud-based commercial APIs, exposing only a truncated output distribution (e.g., top-5 tokens). In contrast, our synthetic data generation approach does not require the teacher's output distribution.

\section{Conclusion, Limitations and Future Work}
In this work, we have elucidated that fine-tuning teacher LLMs to both annotate unlabeled instances and generate new data points can effectively improve a downstream model's performance. Our empirical investigations spanned six tasks, four in classification and two pertaining to natural language generation. 
A potential constraint of our approach is that fine-tuning a large model necessitates significant resources. In future work, we aim to delve deeper into quantifying the fine-tuning required to steer the teacher model towards producing high-quality synthetic data. 
\bibliographystyle{icml}
\bibliography{ref}

\begin{thebibliography}{40}
\providecommand{\natexlab}[1]{#1}
\providecommand{\url}[1]{\texttt{#1}}
\expandafter\ifx\csname urlstyle\endcsname\relax
  \providecommand{\doi}[1]{doi: #1}\else
  \providecommand{\doi}{doi: \begingroup \urlstyle{rm}\Url}\fi

\bibitem[Ope(2023)]{OpenLLMLeaderboarda}
Open {LLM} {Leaderboard} - a {Hugging} {Face} {Space} by {HuggingFaceH4}, 2023.
\newblock URL \url{https://huggingface.co/spaces/HuggingFaceH4/open_llm_leaderboard}.

\bibitem[Andonian et~al.(2021)Andonian, Biderman, Black, Gali, Gao, Hallahan, Levy-Kramer, Leahy, Nestler, Parker, Pieler, Purohit, Songz, Phil, and Weinbach]{Andonian_GPT-NeoX_Large_Scale_2021}
Andonian, A., Biderman, S., Black, S., Gali, P., Gao, L., Hallahan, E., Levy-Kramer, J., Leahy, C., Nestler, L., Parker, K., Pieler, M., Purohit, S., Songz, T., Phil, W., and Weinbach, S.
\newblock {GPT-NeoX: Large Scale Autoregressive Language Modeling in PyTorch}, August 2021.
\newblock URL \url{https://www.github.com/eleutherai/gpt-neox}.

\bibitem[Azerbayev et~al.(2022)Azerbayev, Ni, Schoelkopf, and Radev]{code_generation_knowledge_transfer}
Azerbayev, Z., Ni, A., Schoelkopf, H., and Radev, D.
\newblock Explicit knowledge transfer for weakly-supervised code generation, 2022.
\newblock URL \url{https://arxiv.org/abs/2211.16740}.

\bibitem[Bastianelli et~al.(2020)Bastianelli, Vanzo, Swietojanski, and Rieser]{slurp}
Bastianelli, E., Vanzo, A., Swietojanski, P., and Rieser, V.
\newblock Slurp: A spoken language understanding resource package, 2020.
\newblock URL \url{https://arxiv.org/abs/2011.13205}.

\bibitem[Bayer et~al.(2022)Bayer, Kaufhold, and Reuter]{bayer2022survey}
Bayer, M., Kaufhold, M.-A., and Reuter, C.
\newblock A survey on data augmentation for text classification.
\newblock \emph{ACM Computing Surveys}, 55\penalty0 (7):\penalty0 1--39, 2022.

\bibitem[Black et~al.(2022)Black, Biderman, Hallahan, Anthony, Gao, Golding, He, Leahy, McDonell, Phang, et~al.]{neox}
Black, S., Biderman, S., Hallahan, E., Anthony, Q., Gao, L., Golding, L., He, H., Leahy, C., McDonell, K., Phang, J., et~al.
\newblock Gpt-neox-20b: An open-source autoregressive language model.
\newblock \emph{arXiv preprint arXiv:2204.06745}, 2022.

\bibitem[Bonifacio et~al.(2022)Bonifacio, Abonizio, Fadaee, and Nogueira]{InPars}
Bonifacio, L., Abonizio, H., Fadaee, M., and Nogueira, R.
\newblock Inpars: Data augmentation for information retrieval using large language models, 2022.
\newblock URL \url{https://arxiv.org/abs/2202.05144}.

\bibitem[Brown et~al.(2020)Brown, Mann, Ryder, Subbiah, Kaplan, Dhariwal, Neelakantan, Shyam, Sastry, Askell, et~al.]{gpt3}
Brown, T., Mann, B., Ryder, N., Subbiah, M., Kaplan, J.~D., Dhariwal, P., Neelakantan, A., Shyam, P., Sastry, G., Askell, A., et~al.
\newblock Language models are few-shot learners.
\newblock \emph{Advances in neural information processing systems}, 33:\penalty0 1877--1901, 2020.

\bibitem[Chen et~al.(2022)Chen, Papangelis, Tao, Rosenbaum, Kim, Liu, Yu, and Hakkani-Tur]{chen2022weakly}
Chen, M., Papangelis, A., Tao, C., Rosenbaum, A., Kim, S., Liu, Y., Yu, Z., and Hakkani-Tur, D.
\newblock Weakly supervised data augmentation through prompting for dialogue understanding.
\newblock In \emph{NeurIPS 2022 Workshop on Synthetic Data for Empowering ML Research}, 2022.
\newblock URL \url{https://openreview.net/forum?id=r2_9r7seD-q}.

\bibitem[Clark et~al.(2019)Clark, Lee, Chang, Kwiatkowski, Collins, and Toutanova]{boolq}
Clark, C., Lee, K., Chang, M.-W., Kwiatkowski, T., Collins, M., and Toutanova, K.
\newblock Boolq: Exploring the surprising difficulty of natural yes/no questions.
\newblock \emph{arXiv preprint arXiv:1905.10044}, 2019.

\bibitem[Dai \& Le(2015)Dai and Le]{finetuning_1}
Dai, A.~M. and Le, Q.~V.
\newblock Semi-supervised sequence learning, 2015.
\newblock URL \url{https://arxiv.org/abs/1511.01432}.

\bibitem[Devlin et~al.(2018)Devlin, Chang, Lee, and Toutanova]{bert}
Devlin, J., Chang, M.-W., Lee, K., and Toutanova, K.
\newblock Bert: Pre-training of deep bidirectional transformers for language understanding, 2018.
\newblock URL \url{https://arxiv.org/abs/1810.04805}.

\bibitem[Ding et~al.(2022)Ding, Qin, Liu, Bing, Joty, and Li]{good_annotator}
Ding, B., Qin, C., Liu, L., Bing, L., Joty, S., and Li, B.
\newblock Is gpt-3 a good data annotator?, 2022.
\newblock URL \url{https://arxiv.org/abs/2212.10450}.

\bibitem[Efrat \& Levy(2020)Efrat and Levy]{efrat2020turking}
Efrat, A. and Levy, O.
\newblock The turking test: Can language models understand instructions?
\newblock \emph{arXiv preprint arXiv:2010.11982}, 2020.

\bibitem[Gardent et~al.(2017)Gardent, Shimorina, Narayan, and Perez-Beltrachini]{webnlg}
Gardent, C., Shimorina, A., Narayan, S., and Perez-Beltrachini, L.
\newblock Creating training corpora for nlg micro-planning.
\newblock In \emph{55th annual meeting of the Association for Computational Linguistics (ACL)}, 2017.

\bibitem[Gehrmann et~al.(2021)Gehrmann, Adewumi, Aggarwal, Ammanamanchi, Anuoluwapo, Bosselut, Chandu, Clinciu, Das, Dhole, et~al.]{gem}
Gehrmann, S., Adewumi, T., Aggarwal, K., Ammanamanchi, P.~S., Anuoluwapo, A., Bosselut, A., Chandu, K.~R., Clinciu, M., Das, D., Dhole, K.~D., et~al.
\newblock The gem benchmark: Natural language generation, its evaluation and metrics.
\newblock \emph{arXiv preprint arXiv:2102.01672}, 2021.

\bibitem[Gunasekar et~al.(2023)Gunasekar, Zhang, Aneja, Mendes, Del~Giorno, Gopi, Javaheripi, Kauffmann, de~Rosa, Saarikivi, et~al.]{gunasekar2023textbooks}
Gunasekar, S., Zhang, Y., Aneja, J., Mendes, C. C.~T., Del~Giorno, A., Gopi, S., Javaheripi, M., Kauffmann, P., de~Rosa, G., Saarikivi, O., et~al.
\newblock Textbooks are all you need.
\newblock \emph{arXiv preprint arXiv:2306.11644}, 2023.

\bibitem[Hinton et~al.(2015)Hinton, Vinyals, Dean, et~al.]{hinton2015distilling}
Hinton, G., Vinyals, O., Dean, J., et~al.
\newblock Distilling the knowledge in a neural network.
\newblock \emph{arXiv preprint arXiv:1503.02531}, 2\penalty0 (7), 2015.

\bibitem[Ho et~al.(2022)Ho, Schmid, and Yun]{ho2022large}
Ho, N., Schmid, L., and Yun, S.-Y.
\newblock Large language models are reasoning teachers.
\newblock \emph{arXiv preprint arXiv:2212.10071}, 2022.

\bibitem[Jin et~al.(2019)Jin, Dhingra, Liu, Cohen, and Lu]{jin-etal-2019-pubmedqa}
Jin, Q., Dhingra, B., Liu, Z., Cohen, W., and Lu, X.
\newblock {P}ub{M}ed{QA}: A dataset for biomedical research question answering.
\newblock In \emph{Proceedings of the 2019 Conference on Empirical Methods in Natural Language Processing and the 9th International Joint Conference on Natural Language Processing (EMNLP-IJCNLP)}, pp.\  2567--2577, Hong Kong, China, November 2019. Association for Computational Linguistics.
\newblock \doi{10.18653/v1/D19-1259}.
\newblock URL \url{https://aclanthology.org/D19-1259}.

\bibitem[Kaddour et~al.(2023)Kaddour, Harris, Mozes, Bradley, Raileanu, and McHardy]{callms}
Kaddour, J., Harris, J., Mozes, M., Bradley, H., Raileanu, R., and McHardy, R.
\newblock Challenges and {Applications} of {Large} {Language} {Models}, July 2023.
\newblock URL \url{http://arxiv.org/abs/2307.10169}.
\newblock arXiv:2307.10169 [cs].

\bibitem[Khashabi et~al.(2018)Khashabi, Chaturvedi, Roth, Upadhyay, and Roth]{multirc}
Khashabi, D., Chaturvedi, S., Roth, M., Upadhyay, S., and Roth, D.
\newblock Looking beyond the surface: A challenge set for reading comprehension over multiple sentences.
\newblock In \emph{Proceedings of the 2018 Conference of the North American Chapter of the Association for Computational Linguistics: Human Language Technologies, Volume 1 (Long Papers)}, pp.\  252--262, 2018.

\bibitem[Lewis et~al.(2019)Lewis, Liu, Goyal, Ghazvininejad, Mohamed, Levy, Stoyanov, and Zettlemoyer]{bart}
Lewis, M., Liu, Y., Goyal, N., Ghazvininejad, M., Mohamed, A., Levy, O., Stoyanov, V., and Zettlemoyer, L.
\newblock Bart: Denoising sequence-to-sequence pre-training for natural language generation, translation, and comprehension.
\newblock \emph{arXiv preprint arXiv:1910.13461}, 2019.

\bibitem[Li et~al.(2023)Li, Bubeck, Eldan, Del~Giorno, Gunasekar, and Lee]{li2023textbooks}
Li, Y., Bubeck, S., Eldan, R., Del~Giorno, A., Gunasekar, S., and Lee, Y.~T.
\newblock Textbooks are all you need ii: phi-1.5 technical report.
\newblock \emph{arXiv preprint arXiv:2309.05463}, 2023.

\bibitem[Lin(2004)]{lin2004rouge}
Lin, C.-Y.
\newblock Rouge: A package for automatic evaluation of summaries.
\newblock In \emph{Text summarization branches out}, pp.\  74--81, 2004.

\bibitem[Liu et~al.(2019)Liu, Ott, Goyal, Du, Joshi, Chen, Levy, Lewis, Zettlemoyer, and Stoyanov]{roberta}
Liu, Y., Ott, M., Goyal, N., Du, J., Joshi, M., Chen, D., Levy, O., Lewis, M., Zettlemoyer, L., and Stoyanov, V.
\newblock Roberta: A robustly optimized bert pretraining approach.
\newblock \emph{arXiv preprint arXiv:1907.11692}, 2019.

\bibitem[Meng et~al.(2022)Meng, Huang, Zhang, and Han]{supergen}
Meng, Y., Huang, J., Zhang, Y., and Han, J.
\newblock Generating training data with language models: Towards zero-shot language understanding.
\newblock In Oh, A.~H., Agarwal, A., Belgrave, D., and Cho, K. (eds.), \emph{Advances in Neural Information Processing Systems}, 2022.
\newblock URL \url{https://openreview.net/forum?id=4G1Sfp_1sz7}.

\bibitem[Raffel et~al.(2020)Raffel, Shazeer, Roberts, Lee, Narang, Matena, Zhou, Li, Liu, et~al.]{T5}
Raffel, C., Shazeer, N., Roberts, A., Lee, K., Narang, S., Matena, M., Zhou, Y., Li, W., Liu, P.~J., et~al.
\newblock Exploring the limits of transfer learning with a unified text-to-text transformer.
\newblock \emph{J. Mach. Learn. Res.}, 21\penalty0 (140):\penalty0 1--67, 2020.

\bibitem[Rasley et~al.(2020)Rasley, Rajbhandari, Ruwase, and He]{rasley2020deepspeed}
Rasley, J., Rajbhandari, S., Ruwase, O., and He, Y.
\newblock Deepspeed: System optimizations enable training deep learning models with over 100 billion parameters.
\newblock In \emph{Proceedings of the 26th ACM SIGKDD International Conference on Knowledge Discovery \& Data Mining}, pp.\  3505--3506, 2020.

\bibitem[Rastogi et~al.(2019)Rastogi, Zang, Sunkara, Gupta, and Khaitan]{sgd}
Rastogi, A., Zang, X., Sunkara, S., Gupta, R., and Khaitan, P.
\newblock Towards scalable multi-domain conversational agents: The schema-guided dialogue dataset, 2019.
\newblock URL \url{https://arxiv.org/abs/1909.05855}.

\bibitem[Sahu et~al.(2022)Sahu, Rodriguez, Laradji, Atighehchian, Vazquez, and Bahdanau]{sahu2022data}
Sahu, G., Rodriguez, P., Laradji, I.~H., Atighehchian, P., Vazquez, D., and Bahdanau, D.
\newblock Data augmentation for intent classification with off-the-shelf large language models.
\newblock \emph{arXiv preprint arXiv:2204.01959}, 2022.

\bibitem[Schick \& Schütze(2021)Schick and Schütze]{dino}
Schick, T. and Schütze, H.
\newblock Generating datasets with pretrained language models, 2021.
\newblock URL \url{https://arxiv.org/abs/2104.07540}.

\bibitem[Wang et~al.(2018)Wang, Singh, Michael, Hill, Levy, and Bowman]{glue}
Wang, A., Singh, A., Michael, J., Hill, F., Levy, O., and Bowman, S.
\newblock {GLUE}: A multi-task benchmark and analysis platform for natural language understanding.
\newblock In \emph{Proceedings of the 2018 {EMNLP} Workshop {B}lackbox{NLP}: Analyzing and Interpreting Neural Networks for {NLP}}, pp.\  353--355, Brussels, Belgium, November 2018. Association for Computational Linguistics.
\newblock \doi{10.18653/v1/W18-5446}.
\newblock URL \url{https://aclanthology.org/W18-5446}.

\bibitem[Wang et~al.(2019)Wang, Pruksachatkun, Nangia, Singh, Michael, Hill, Levy, and Bowman]{superglue}
Wang, A., Pruksachatkun, Y., Nangia, N., Singh, A., Michael, J., Hill, F., Levy, O., and Bowman, S.
\newblock Superglue: A stickier benchmark for general-purpose language understanding systems.
\newblock \emph{Advances in neural information processing systems}, 32, 2019.

\bibitem[Wang et~al.(2021)Wang, Liu, Xu, Zhu, and Zeng]{reduce_label_cost}
Wang, S., Liu, Y., Xu, Y., Zhu, C., and Zeng, M.
\newblock Want to reduce labeling cost? gpt-3 can help, 2021.
\newblock URL \url{https://arxiv.org/abs/2108.13487}.

\bibitem[Wang et~al.(2022)Wang, Mishra, Alipoormolabashi, Kordi, Mirzaei, Arunkumar, Ashok, Dhanasekaran, Naik, Stap, Pathak, Karamanolakis, Lai, Purohit, Mondal, Anderson, Kuznia, Doshi, Patel, Pal, Moradshahi, Parmar, Purohit, Varshney, Kaza, Verma, Puri, Karia, Sampat, Doshi, Mishra, Reddy, Patro, Dixit, Shen, Baral, Choi, Smith, Hajishirzi, and Khashabi]{wang2022supernaturalinstructions}
Wang, Y., Mishra, S., Alipoormolabashi, P., Kordi, Y., Mirzaei, A., Arunkumar, A., Ashok, A., Dhanasekaran, A.~S., Naik, A., Stap, D., Pathak, E., Karamanolakis, G., Lai, H.~G., Purohit, I., Mondal, I., Anderson, J., Kuznia, K., Doshi, K., Patel, M., Pal, K.~K., Moradshahi, M., Parmar, M., Purohit, M., Varshney, N., Kaza, P.~R., Verma, P., Puri, R.~S., Karia, R., Sampat, S.~K., Doshi, S., Mishra, S., Reddy, S., Patro, S., Dixit, T., Shen, X., Baral, C., Choi, Y., Smith, N.~A., Hajishirzi, H., and Khashabi, D.
\newblock Super-naturalinstructions: Generalization via declarative instructions on 1600+ nlp tasks, 2022.

\bibitem[Wei et~al.(2022)Wei, Tay, Bommasani, Raffel, Zoph, Borgeaud, Yogatama, Bosma, Zhou, Metzler, et~al.]{wei2022emergent}
Wei, J., Tay, Y., Bommasani, R., Raffel, C., Zoph, B., Borgeaud, S., Yogatama, D., Bosma, M., Zhou, D., Metzler, D., et~al.
\newblock Emergent abilities of large language models.
\newblock \emph{arXiv preprint arXiv:2206.07682}, 2022.

\bibitem[Yoo et~al.(2021)Yoo, Park, Kang, Lee, and Park]{yoo2021gpt3mix}
Yoo, K.~M., Park, D., Kang, J., Lee, S.-W., and Park, W.
\newblock Gpt3mix: Leveraging large-scale language models for text augmentation.
\newblock \emph{arXiv preprint arXiv:2104.08826}, 2021.

\bibitem[Zhang et~al.(2021)Zhang, Wu, Katiyar, Weinberger, and Artzi]{zhangRevisitingFewsampleBERT2021b}
Zhang, T., Wu, F., Katiyar, A., Weinberger, K.~Q., and Artzi, Y.
\newblock Revisiting {Few}-sample {BERT} {Fine}-tuning, March 2021.
\newblock URL \url{http://arxiv.org/abs/2006.05987}.
\newblock arXiv:2006.05987 [cs].

\bibitem[Zheng et~al.(2023)Zheng, Sabour, Wen, Zhang, and Huang]{zheng2023augesc}
Zheng, C., Sabour, S., Wen, J., Zhang, Z., and Huang, M.
\newblock Augesc: Dialogue augmentation with large language models for emotional support conversation, 2023.

\end{thebibliography}
\appendix
\section{Exemplary Generations}\label{app:examples}
\onecolumn
\begin{xltabular}{\textwidth}{|X|}
\caption{\textbf{Examples of annotations for WebNLG \cite{webnlg}}. \colorG{Ground-truth references} are shown in the \{braces\}. }
   \label{tab:ex_webnlg_ann}  \\
   \toprule
            \textbf{Input}: \texttt{[CONTEXT]} WrittenWork \texttt{[DATA]} A\_Glastonbury\_Romance | mediaType | Hardcover A\_Glastonbury\_Romance | numberOfPages | 1174 | isbnNumber | 0-7156-3648-0 \\
             \textbf{Output}: A Glastonbury Romance, with 1144 pages, is available in print and is ISBN number 0-7156-3648.
             \\ \colorG{\{A Glastonbury Romance has the ISBN number 0-7156-3648-0, can be found in hardcover and has 1174 pages.\}}\\
            \midrule
            \textbf{Input}: \texttt{[CONTEXT]} MeanOfTransportation \texttt{[DATA]} A-Rosa\_Luna | builder | Germany MTU\_Friedrichshafen | city | Friedrichshafen A-Rosa\_Luna | powerType | MTU\_Friedrichshafen \\
             \textbf{Output}: A-Rosa Luna was designed by MTU Friedrichshafen and is powered by them. The city of Friedrichshafen is where the builder of the A-Rosa Luna is located.
             \\ \colorG{\{The A-Rosa Luna is powered by a MTU Friedrichshafen engine in the city of Friedrichshafen, Germany.\}}\\            \midrule
            \textbf{Input}: \texttt{[CONTEXT]} Astronaut \texttt{[DATA]} Alan\_Bean | nationality | United\_States Alan\_Bean | occupation | Test\_pilot Alan\_Bean | almaMater | UT Austin, B.S. 1955 Alan\_Bean | mission | Apollo\_12 Alan\_Bean | birthPlace | Wheeler,\_Texas \\
             \textbf{Output}: Alan Bean was a national of the United States, who was a student at the University of Texas at Austin (B.S. 1955). He was an American test pilot and was a member of the Apollo 12 crew. Mr. Bean was born in Wheeler, Texas.
             \\ \colorG{\{Alan Bean is a United States citizen born in Wheeler Texas. He served as a test pilot and was a crew member of the Apollo 12 mission.\}}
             \\            \bottomrule
            
\end{xltabular}

\begin{xltabular}{\textwidth}{|X|}
\caption{\textbf{Examples of generations for WebNLG \cite{webnlg}}. }
    \label{tab:ex_webnlg_gen} \\
              \toprule
            \textbf{Input}: \texttt{[CONTEXT]} Astronaut \texttt{[DATA]} Elliot\_See | almaMater | University\_of\_Texas\_at\_Austin University\_of\_Texas\_at\_Austin | affiliation | University\_of\_Texas\_System Elliot\_See | occupation | Test\_pilot Elliot\_See | nationality | United\_States Elliot\_See | birthPlace | Dallas Elliot\_See | deathPlace | St.\_Louis Elliot\_See | status | Deceased \\
             \textbf{Output}: Elliot See was a US national, who was born in Dallas. He was a student at the University of Texas at Austin (affiliated to the university of Texas system). He was a test pilot who died in St. Louis.
             \\ 
            \midrule
            \textbf{Input}: \texttt{[CONTEXT]} Airport \texttt{[DATA]} Al\_Asad\_Airbase | operatingOrganisation | United\_States\_Air\_Force United\_States\_Air\_Force | transportAircraft | Lockheed\_C-130\_Hercules \\
             \textbf{Output}: Al Asad Airbase is operated by the United States Air Force which use the Lockheed C-130 Hercules as a transport aircraft.
             \\ 
            \midrule
            \textbf{Input}: \texttt{[CONTEXT]} Food \texttt{[DATA]} Ajoblanco | ingredient | Almond Spain | leader | Felipe\_VI\_of\_Spain Ajoblanco | region | Andalusia Andalusia | leader | Susana\_D\u00edaz Ajoblanco | country | Spain \\
             \textbf{Output}: Ajoblanco originates from Spain where Felipe VI is leader. Ajoblanco is found in Andalusia where Susana Diaz is leader. One of the ingredients of ajoblanco is almonds.
             \\ \bottomrule            
\end{xltabular}

\begin{xltabular}{\textwidth}{|X|}
            \caption{\textbf{Examples of annotations for SGD \cite{sgd}}. \colorG{Ground-truth references} are shown in the \{braces\}. } \label{tab:ex_sgd_ann}\\
          \toprule
            \textbf{Input}: \texttt{[CONTEXT]} User: I want to make the most of the free time I have coming up by finding something fun to do. Let's say I have no idea what to do but my trusty friend, that's you, will help me. Am I right? \\
             \textbf{Output}: I'll be happy to help. What type of events do you have in mind?
             \\ \colorG{\{Sure, I'll be happy to help you with that. Let's start with this: tell me what type of activity and the area you're interested in.\}}\\
            \midrule
            \textbf{Input}: \texttt{[CONTEXT]} User: Can you find me oneway flights? \\ System: Where will you be flying from? \\User: I am flying out of LA. \\ System: When do you want to leave? \\ User: I need to leave on the 10th of this month. \\ System: Where will you be flying too. \\ User: I am flying to Paris, France with 4 people. I prefer Delta Airlines. \\
             \textbf{Output}: Please confirm the following: You will need 1 Economy ticket on Delta Airlines. You will be flying from Los Angeles to Paris on March 10th. You will be traveling with 4 people.
             \\ \colorG{\{There is a direct Delta Airlines flight that leaves at 8:25 am and has 0 stops for \$255.\}}\\
            \midrule
            \textbf{Input}: \texttt{[CONTEXT]} User: I'd like to find a rental car to pick up in New York.\\ System: When will you need it?\\ User: I would like to pick it up at 12 in the afternoon.\\ System: What date would you like to return the car?\\ User: I'd like to return it on the 9th of March.\\ System: What date would you like to pick it up?\\ User: I want to pick it up Friday next week.\\ System: I found a standard size Honda Accord for you that's available at Grand Central Station on March 8th, does that work?\\ User: Is there anything else available? I would like a compact size and will actually need it until March 10th.\\ System: Sure, I found a compact size Chevrolet Bolt that is available at Port Authority Station on March 8th, does that work for you?\\ User: What's the cost for the car?\\ System: The total cost is \$56.\\ User: Is there anything available in Paris, France? \\
             \textbf{Output}: There are 10 cars available. How about a compact size Chevrolet Cruze? It's available to pick up from Charles de Gaulle International Airport on March 8th.
             \\ \colorG{\{Yes, there is a compact size Honda Fit available at the Charles de Gaulle International Airport on March 8th.\}}\\
            \bottomrule
    \end{xltabular}
    
    \begin{xltabular}{\textwidth}{|X|}
        \caption{\textbf{Examples of generations for SGD \cite{sgd}}.}     \label{tab:ex_sgd_gen} \\
            \toprule
            \textbf{Input}: \texttt{[CONTEXT]} User: I want to eat at a place in San Francisco.\\ System: What type of food? You can say Indian, Italian, Vegetarian, or something else.\\ User: Small plates would be perfect.\\ System: Aatxe has small plates and is in San Francisco.\\ User: What's the address and is their live music?\\ System: Their address is 2174 Market Street and they don't have live music.\\ User: What else is available? Check Redwood City.\\ System: La Viga Seafood \& Cocina Mexicana is in Redwood City.\\ User: Yes, that will work.\\ System: Do you want to book a table?\\ User: Not now. Can you find a salon in Redwood City?\\ System: August Barbershop is in Redwood City.\\ User: Yes, I've heard of them.\\ System: Shall I make an appointment.\\ User: Please do, book it for half past 2 in the afternoon.\\ System: What date do you want it for?\\ User: The 4th of March and book it for two pm.\\ System: Please confirm the following: Booking an appointment with August Barbershop for March 4th at 2 pm.\\ User: No, do it for the 1st of March.\\ System: Please confirm the following: Booking an appointment with August Barbershop for March 1st.\\ User: Yes, that's correct.\\ System: Your appointment went through successfully!\\ User: Could you try booking a table at that La Viga place?\\ System: What time is it for?\\ User: It's for half past 6 in the evening.\\ System: Please confirm the following: Booking a table for 2 at La Viga Seafood \& Cocina Mexicana in Redwood City on March 1st at 6:30 pm.\\ User: Yes, that's correct.\\ System: Your appointment went through successfully!\\ User: Could you try booking a table for 4 on the 7th of this month?\\ System: Please confirm the following: Booking a table for 4 at La Viga Seafood \& Cocina Mexicana in Redwood City on March 7th.\\ User: Yes, that's correct.\\ System: Your appointment went through successfully!\\ User: Thanks a lot. \\
             \textbf{Output}: Do you need anything else?
             \\ 
            \midrule
            \textbf{Input}: \texttt{[CONTEXT]} User: I'm looking for some events.\\System: What city should I look in?\\User: I want to see something in New York City.\\System: What kind of events are you looking for?\\User: I want to see Carmina Burana.\\System: I found 2 occurrences of the event Carmina Burana. The first is at the Sony Hall on March 5th.\\User: What is the address of the venue?\\System: The venue is located at 235 West 46th Street.\\User: What is the venue's address?\\System: The address is 235 West 46th Street.\\User: That sounds great.\\System: Do you want to purchase tickets?\\User: No, I don't want to buy tickets right now.\\System: Is there anything else I can help you with?\\User: I want to buy bus tickets to get there.\\System: How many tickets do you need? What time do you want to depart?\\User: I need Economy tickets for 4 people. I want to leave at one pm.\\System: Where will you be departing from?\\User: I'm leaving from Washington.\\System: Okay, I have 4 Economy bus tickets leaving from Washington and going to New York on March 5th at 1 pm. Is that correct?\\User: Yes, that sounds good. What station does the bus depart from and what station does it arrive at? \\
             \textbf{Output}: I've purchased the tickets. The bus leaves from Union Station and arrives at Port Authority Station.
             \\ 
            \midrule
            \textbf{Input}: \texttt{[CONTEXT]} I'm looking for a rental car.\\System: What day and time do you want to pick it up? How long will you be needing it?\\User: I need it at three pm on the 8th of March, until the 11th of March.\\System: Where do you want to pick it up?\\User: I want to get it in Anaheim, CA.\\System: I have 5 cars you might like. How about a Standard Chevrolet Cruze? It's available at John Wayne Airport on March 8th.\\User: What else do you have?\\System: I have a Standard Honda Accord at John Wayne Airport on March 8th.\\User: What else do you have?\\System: I have a Standard Toyota Camry at Anaheim Intermodal Center on March 8th.\\User: That one sounds good.\\System: Do you want to reserve it?\\User: Yes, please reserve it.\\System: Please confirm the following: You are renting a Standard car from Anaheim Intermodal Center on March 8th at 3 pm. You'll return it on March 11th.\\User: Actually, I want to keep it until March 12th.\\System: Please confirm the following details: You are keeping the car until March 12th.\\User: That's correct. How much does it cost?\\System: Your car has been reserved. The total cost is \$170.\\User: I want to find a movie near there.\\System: Do you want to look for movies showing in Anaheim?\\User: Yes, that's what I want.\\System: How about some movies by Cameron Crowe? There's also dogman, which is a drama.\\User: Is there anything else? I want something more family-friendly.\\System: There's also Family, which is a comedy.\\User: Family sounds good.\\System: Do you want to buy tickets?\\User: No, not right now. \\
             \textbf{Output}: Is there anything else you need?
            \\ \bottomrule
\end{xltabular}
\twocolumn


\end{document}